\documentclass{article}

\usepackage[final]{neurips_2022}

\usepackage[utf8]{inputenc} 
\usepackage[T1]{fontenc}    
\usepackage{url}            
\usepackage{booktabs}       
\usepackage{amsfonts}       
\usepackage{nicefrac}       
\usepackage{microtype}      
\usepackage{xcolor}         
\usepackage{graphicx}
\usepackage{amsmath,bm}
\usepackage{subcaption}
\usepackage{wrapfig, blindtext}
\usepackage{lipsum}
\usepackage{relsize}
\usepackage{enumitem}       
\usepackage{todonotes}
\usepackage{amssymb}

\usepackage{caption}
\usepackage{subcaption}

\usepackage{color}
\usepackage{tcolorbox}
\usepackage{CJK}
\usepackage{adjustbox}
\tcbset{width=0.9\textwidth,boxrule=0pt,colback=red, arc=0pt, auto outer arc,left=0pt,right=0pt,boxsep=5pt}
\definecolor{bbe}{rgb}{0.63, 0.79, 0.95}
\definecolor{cb}{rgb}{0.6, 0.73, 0.89}
\definecolor{ceil}{rgb}{0.57, 0.63, 0.81}
\definecolor{celadon}{rgb}{0.67, 0.88, 0.69}
\definecolor{cadmiumorange}{rgb}{0.93, 0.53, 0.18}
\definecolor{brightlavender}{rgb}{0.75, 0.58, 0.89}
\definecolor{asparagus}{rgb}{0.53, 0.66, 0.42}
\definecolor{azure}{rgb}{0.0, 0.5, 1.0}
\definecolor{bleudefrance}{rgb}{0.19, 0.55, 0.91}

\usepackage[pagebackref=true]{hyperref}        
\renewcommand*\backref[1]{\ifx#1\relax \else (Cited on #1) \fi}
 
\DeclareMathOperator*{\argmax}{argmax}

\newcommand{\sm}{\scalebox{0.7}[1.0]{\( - \)}}

\title{Learning to Drop Out: An Adversarial Approach to Training Sequence VAEs}
  
\author{%
  {\DJ}or{\dj}e Miladinovi{\'c} $^{* \dag}$  \quad Kumar Shridhar \thanks{ Equal contribution; correspondence at: shkumar@ethz.ch}\ $\ ^ \ddag$   \quad Kushal Jain $^\mathparagraph$ \\ 
  \bf{Max B. Paulus $^	\ddag$ \quad Joachim M. Buhmann $^	\ddag$ \quad Mrinmaya Sachan $^	\ddag$ \quad Carl Allen $^	\ddag$} \\ \\
  $^{\dag}$GSK.ai  \quad $^\ddag$ETH Z{\"u}rich \quad $^\mathparagraph$University of California, San Diego
}

\begin{document}

\maketitle

\begin{abstract}

In principle, applying variational autoencoders (VAEs) to sequential data offers a method for controlled sequence generation, manipulation, and structured representation learning.
However, training sequence VAEs is challenging: autoregressive decoders can often explain the data without utilizing the latent space, known as \emph{posterior collapse}.
To mitigate this, state-of-the-art models `weaken' the `powerful' decoder by applying uniformly random \textit{dropout} to the decoder input.
We show theoretically that this removes \textit{pointwise mutual information} provided by the decoder input, which is compensated for by utilizing the latent space. We then propose an \emph{adversarial} training strategy to achieve \emph{information-based stochastic dropout}.
Compared to uniform dropout on standard text benchmark datasets, our targeted approach increases both sequence modeling performance and the information captured in the latent space.

\end{abstract}

\section{Introduction}

Training autoregressive models via maximum likelihood
estimation (MLE) is a common strategy for representing sequential data. Such autoregressive models obtain state-of-the-art in modeling text \citep{brown2020language, chowdhery2022palm, zhang2022opt}, speech \citep{oord2018parallel, conneau2020unsupervised} , and video sequences \citep{babaeizadeh2018stochastic}. 
In this simple and intuitive approach, the joint distribution of a sequence is factorized into a product of conditional distributions, with each sequence element conditioned on its history. 
However, in their basic form, autoregressive models do not necessarily learn \textit{latent variables} that encode informative content, as often desired. As such, supplementing autoregressive models with latent variables is a promising way to enhance control in sequence generation, and enable structured
representation learning.

A recent approach to latent variable modeling of sequences is to integrate autoregressive components into the \textit{variational autoencoder} (VAE) framework \citep{kingma2013auto, rezende2014stochastic}.
\emph{Sequence VAEs} offer a theoretically principled solution, but have not been widely adopted. Arguably, this is largely due to the phenomenon of \emph{posterior collapse}  \citep{bowman2016,chen2016variational,van2017neural, li2020improving}. Posterior collapse describes when a VAE's posterior probability over latent variables `collapses' to the latent prior, rendering the latent space completely uninformative, or \textit{independent}, of the data.
Recent years have seen an abundance of research into posterior collapse \citep{bowman-etal-2015-large,chen2016variational,lucas2019don,dai2020usual, Fang_Bai_Xu_Lyu_King_2020, Pang_Nijkamp_Han_Wu_2021}, identifying the `power' of autoregressive decoding as a main cause.
In particular, autoregressive decoders are shown to obtain satisfactory sequence modeling performance without utilizing the latent space \citep{bowman2016,chen2016variational}.
This suggests that for sequence VAEs to progress, improved techniques are needed to understand and alleviate posterior collapse, and thereby learn informative latent structure.
In this work, we propose a theoretically principled way to mitigate posterior collapse when training autoregressive sequence VAEs.

\begin{figure*}[!t] 
\centering	
\includegraphics[width=\textwidth]{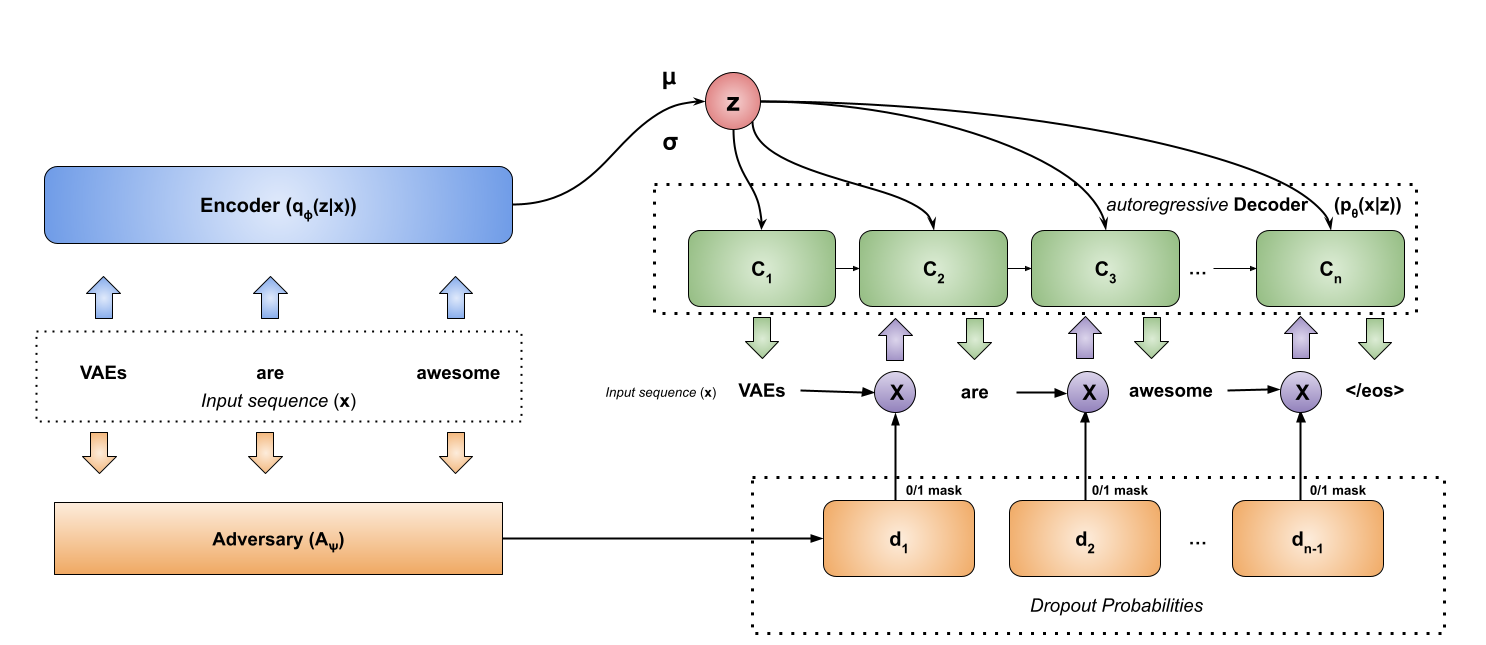}
\caption{\text{Adversarial training of sequence VAEs.} Our proposed model comprises {\color{bleudefrance}\textbf{an encoder}}, {\color{asparagus}\textbf{an autoregressive sequence decoder}} (together an autoregressive VAE), and  {\color{cadmiumorange}\textbf{an adversary A$_{\bm{\psi}}$}}. 
During training, the encoder learns a representation of a full input sequence $\bm{x}$; the same sequence serves as both input (together with the hidden representation) and target to train the decoder. The \text{adversary} \emph{learns to stochastically drop out} sequence elements that the decoder requires most, {\color{brightlavender}\textbf{masking (0/1)}} each sequence element $\bm{x}_i$ with {\color{cadmiumorange}\textbf{probability $\bm{d_i}$}}. See Section \ref{sec:method} for details.}
\label{fig-adversarial-dropout}
\vspace{-8pt}
\end{figure*}

The intuition that autoregressive VAE decoders are too `powerful' has led to the notion of `weakening' them, e.g.\ by regularization. However, weakening an autoregressive model arbitrarily might of course harm its performance, leading to a trade off between informativeness of the latent space and the quality of sequence modeling (or density estimation performance).
State-of-the-art VAEs apply variants of \textit{dropout} \citep{srivastava2014dropout} to the sequence input into the decoder: either to individual dimensions of sequence elements \citep{kim2018semi}, or to mask entire sequence elements \citep{iyyer2015deep,bowman2016}.
In either case, dropout is generally applied uniformly at random, meaning that each sequence element has an equal probability of being dropped.
From the intuition that \text{different words carry different information} about the next word, this work investigates whether a \textit{non-uniform} dropout policy can achieve a better trade-off between capturing latent information and sequence modeling performance. 
Specifically, by stochastically dropping each sequence element according to its importance to the autoregressive decoder, our method takes a \textit{targeted} approach to dropout relative to uniform sampling, maintaining sequence modeling performance while achieving the benefits of dropout on the latent space.
To apply a non-uniform, data-dependent dropout scheme, our approach introduces an `adversary' that \emph{learns which elements to drop}. 
The proposed framework is represented in Figure~\ref{fig-adversarial-dropout}, theoretically developed in Section \ref{sec:awd_theory} and implemented in Section \ref{sec:method}.

The contributions of this paper can be summarized as follows: 
\ \textbf{(i)} We propose an adversarial approach to learning a stochastic dropout policy that mitigates posterior collapse in sequence VAEs.
\ \textbf{(ii)} We show theoretically that dropping out sequence elements \text{deducts} \textit{mutual information not already learned by the model}, between one sequence element and the next, and that our approach maximises this quantity, leading to a constrained minimax ELBO objective that explains an adversarial approach.
\ \textbf{(iii)} We evaluate the proposed scheme on standard text benchmarks, showing that our approach renders an informative latent space, without trading-off but rather improving sentence modeling. 
\ \textbf{(iv)} We examine properties of the adversarial network and find that the adversary typically selects words that carry sentence semantics that fit our theoretical analysis.

\section{Background} \label{sec:bg}

Here, we formally describe the use of autoregressive decoding within VAEs for sequential data and the problem of \textit{posterior collapse} that often arises (refer to Appendix \ref{sec-posterior-def} for some clarification around the term \textit{posterior collapse}).

For notation:
    $\bm{X}$ denotes a random variable, 
    (boldface) $\bm{x}$ denotes a \textit{sequence} that is a realization of $\bm{X}$,
    (subscript) $\bm{x}_i$ denotes the $i$-th sequence element of $\bm{x}$ and
     $\bm{x}_{<i}$ collectively denotes all sequence elements prior $\bm{x}_i$. 
    $z$ denotes a global latent vector and (superscript) $\bm{x}^{(i)}$ denotes the $i$-th sample from a dataset.
    

The task is to approximate the true probability density function $p_*(\bm{x})$ of a random sequence $\bm{X} \sim p_*(\bm{x})$ using a model $p_\theta(\bm{x})$ with parameters $\theta$. Given a set of observations $\{\bm{x}^{(i)}\}^N_{i=1}$, $\theta$ can be estimated by \textit{maximum likelihood estimation} (MLE):
\begin{align}
\label{eq:mle}
\theta_{\emph{MLE}} = \argmax_{\theta} \left[\frac{1}{N} \sum_{i=1}^N \log p_{\theta}(\bm{x}^{(i)}) \right]
\end{align}

If the data has latent generative factors $z$, modelled as $p_\theta(\bm{x}) \!=\! \int_zp_\theta(\bm{x}|z)p_\theta(z)$, Eq~\ref{eq:mle} is intractable and one can instead maximize the \textit{evidence lower bound} (ELBO), which for any observation $\bm{x}$ is:

\begin{align}
    \label{eq:elbo}
    \log p(\bm{x}) 
        \,\geq \log p_\theta(\bm{x}) 
        \,\geq \text{ELBO}_{\theta, \phi}(\bm{x}) 
        \,\triangleq\! \int_z q_{\phi}(z|\bm{x}) \log p_\theta (\bm{x}|z) 
        - \int_z q_{\phi}(z|\bm{x})\log\tfrac{q_{\phi}(z|\bm{x})}{p_\theta(z)} 
\end{align}
where, $q_{\phi}(z|\bm{x})$ approximates the posterior, parameterized by ${\phi}$, and $p_\theta(z)$ is a prior distribution over latent variables.
In this work, we consider the framework of variational autoencoders (VAEs) \citep{kingma2013auto, rezende2014stochastic}, where the \textit{encoder} $q_\phi(z|\bm{x})$ and \textit{decoder} $p_\theta(\bm{x}|z)$ are parameterised by neural networks and $p_\theta(z)$ is a standard Gaussian, denoted $p_0(z)$.
Further, we focus specifically on VAEs where the \text{decoder} $p_{\theta}(\bm{x}|z)$ is \textit{autoregressive} \citep{bowman2016, li2020improving, he2018lagging}, meaning 
    $\log p_{\theta}(\bm{x}|z) \!=\! \sum_i \log p_{\theta}({x}_i|\bm{x}_{<i},z)$, and the ELBO becomes:

\begin{flalign} 
    &&\text{ELBO}^{{\textbf{AR}}}_{\theta, \phi}(\bm{x})  
      &= \int_z q_{\phi}(z|\bm{x}) \sum_i \log p_\theta (\bm{x}_i|\bm{x}_{<i}, z) 
        - \underbrace{\int_z q_{\phi}(z|\bm{x})\log\tfrac{q_{\phi}(z|\bm{x})}{p_0(z)}}_{\text{KL}} &&
    \label{eq:elbo_AR}
\end{flalign}
\paragraph{Posterior collapse:} Autoregressive models improve density estimation \citep{oord2016conditional} by explicitly modeling statistical dependencies between sequence elements.
However, when used as a VAE decoder $p_\theta(\bm{x}|z)$, autoregressive models are often found to ignore $z$, i.e.\ $p(\bm{x}|z)\!\approx\! p(\bm{x})$, and the posterior is said to `collapse' to the prior, i.e.\ $q_\theta(z|\bm{x})\!\approx\! p_0(z)$, rendering $z$ \emph{de facto} independent of $\bm{x}$ \citep{bowman2016, he2018lagging}. 
In terms of Eq~\ref{eq:elbo_AR}, this implies $p_{\theta}(\bm{x}_i| \bm{x}_{<i},z) \!\approx\! p_{\theta}(\bm{x}_i|\bm{x}_{<i})$.
From this, posterior collapse can be considered due to the \textit{information} about $\bm{x}_i$ provided by $\bm{x}_{<i}$ being such that each $\bm{x}_i$ is (approximately) conditionally independent of $z$ given $\bm{x}_{<i}$, in other words $z$ is essentially \textit{redundant}. Based on this, we interpret the notion that autoregressive models are \textit{too powerful} for use as a VAE decoder to mean that they allow \textit{too much information flow}.

\section{Related Work}
\label{sec:related-work}

\paragraph{Posterior collapse}
The phenomenon of posterior collapse has been observed in the context of text \citep{bowman2016,yang2017improved}, images \citep{chen2016variational,razavi2018preventing,miladinovic2021spatial}, videos \citep{babaeizadeh2018stochastic,miladinovic2019disentangled}, speech \citep{chorowski2019unsupervised} and graphs \citep{kipf2018neural}.
In broad terms, previous solutions can be divided into two complementary categories: 
\textbf{(i)} \emph{latent-variable-oriented} methods that focus mainly on relaxing the KL penalty; and 
\textbf{(ii)} \emph{decoder-oriented} methods that regularise or `weaken' the autoregressive decoder.
Typically, both types are required in order to learn an informative latent space \citep{bowman2016,chen2016variational,kim2018semi}. 

\textbf{Latent-variable-oriented solutions}
\cite{bowman2016} suggest \emph{KL annealing}, a technique that introduces the KL term gradually into training according to a predefined schedule.
\emph{Free bits} \citep{kingma2016improved} prevent penalizing of the KL term if its magnitude is below a predefined threshold.
\cite{van2017neural} use a \textit{discrete} VAE, avoiding posterior collapse by design.
\cite{fu-etal-2019-cyclical} apply a \emph{cyclical} form of KL annealing.
\cite{razavi2018preventing} constrain the variational family of the posterior distribution (encoder), preventing it from closely approximating the prior and so holding the KL term away from zero.
\emph{Lagging inference networks} \citep{he2018lagging} `aggressively' optimize the encoder before each decoder update. A similar procedure is followed by \emph{Semi-Amortized VAEs} \citep{kim2018semi}.
\emph{Generative skip models} \citep{dieng2019avoiding} introduce skip connections to create a more explicit link between the latent variables and the likelihood function.
\cite{NEURIPS2021_6c19e0a6} propose a \textit{consistency} regularizer by minimizing the KL divergence between the posterior approximations of an observation and a random transformation of it.

\textbf{Decoder-oriented solutions}
\cite{bowman2016} apply \emph{word dropout} \citep{iyyer2015deep} to uniformly drop words during autoregressive decoding.
Other methods \citep{kim2018semi,he2018lagging} apply parameter dropout \citep{srivastava2014dropout} to word embeddings.
\cite{chen2016variational,semeniuta-etal-2017-hybrid,yang2017improved} constrain the receptive field of the decoder, limiting the window of autoregression, however, this is not readily applicable to RNN architectures given their unbounded receptive field.
Our proposed adversarial method falls into this category as an extension of word dropout, which is then subsumed as a special case of adversarial word dropout (Section \ref{sec:method}).

\paragraph{Non-uniform dropout}
Previous work has recognized the benefit of adapting the dropout rate across different architectural components, though in entirely different contexts \citep{kingma2015variational,gal2017concrete,achille2018information}.
These works introduce different approaches to regularizing the magnitude of weights or activations (groups of weights). Elements that are deemed \emph{irrelevant} during training are then dropped. For instance, variational dropout \citep{kingma2015variational} has been used to prune weights in deep neural networks \citep{molchanov2017variational}. A key difference to our approach is that instead of a regularization term, we employ an adversary that selects elements for dropout during training based on their information content not yet learned by the model.

\paragraph{Exposure bias}
The method by which the autoregressive decoder is trained, known as \textit{teacher forcing}, bears a connection to \emph{exposure bias}, which refers to the gap between training and inference \citep{ranzato2015sequence}.
Namely, in language generation, a trained model generates sequences at test time without access to the ground truth history that was accessible during training. Since the model was not trained to continue its predictions, this can lead to error accumulation \citep{ranzato2015sequence}.
The most common approach to tackle exposure bias is to switch between conditioning on ground truth and model predictions,  with the latter being preferred towards the end of training \citep{daume2009search,bengio2015scheduled,ranzato2015sequence}. 
In preliminary studies, we implemented \emph{scheduled sampling} \citep{bengio2015scheduled} but did not find it to outperform even uniform dropout.
\section{Adversarial Word Dropout (AWD)}

\subsection{Theoretical Basis}
\label{sec:awd_theory}

As a precursor to our adversarial approach, we first derive the effect of stochastic dropout on the ELBO under the autoregressive assumption (Equation~\ref{eq:elbo_AR}).

\textbf{Word dropout:} Our interpretation of posterior collapse at the end of Section \ref{sec:bg} suggests that to `weaken the decoder' one should \textit{restrict the available information}, such that $z$ is no longer redundant and must capture some of the restricted information. Word dropout \citep{bowman2016} can be seen to do this by stochastically masking the previous word $\bm{x}_{i\sm1}$ with probability $d_i$, which substitutes $p_{\theta} (\bm{x}_i|\bm{x}_{{<i}},z)$ with $p_{\theta} (\bm{x}_i|\bm{x}_{{<i\sm1}},z)$ in Equation \ref{eq:elbo_AR}, i.e.:
%
\begin{flalign} 
    \label{eq:elbo_DO_flat}
    \text{ELBO}^{{\textbf{WD}}}_{\theta, \phi}(\bm{x})  
    &= \int_z q_{\phi}(z|\bm{x})
        \Big\{\sum_i (1\,\sm\, d_i)\log p_{\theta} (\bm{x}_i|\bm{x}_{{<i}},z) + d_i\log p_{\theta} (\bm{x}_i|\bm{x}_{{<i\sm1}},z)\Big\}
        -\text{KL} &&
    \\
    \label{eq:elbo_DO}
     &= 
    \int_z q_{\phi}(z|\bm{x})\Big\{\sum_i
        \log p_{\theta} (\bm{x}_i|\bm{x}_{\tiny{<i}},z)
            -  \sum_i d_i\ 
            \!\! \underbrace{ \log
            \tfrac{p_{\theta} (\bm{x}_i|\bm{x}_{\tiny{i\sm1}}, \bm{x}_{\tiny{<i\sm1}},z)}{
                    p_{\theta} (\bm{x}_i|\bm{x}_{\tiny <i\sm1},z)}
                }_{\text{PMI}(\bm{x}_i,\bm{x}_{i-1}|\bm{x}_{<i-1},z)}
                \Big\}
        -\text{KL} &&
\end{flalign}
Equation \ref{eq:elbo_DO} shows the effect (in expectation) of applying word dropout when maximising the ELBO with an autoregressive decoder in Eq~\ref{eq:elbo_AR} (of which `KL' denotes the last term). In effect, a weighted sum is introduced with weights given by dropout probabilities $d_i$ and components that define conditional \textit{point-wise mutual information} (PMI), where classic point-wise mutual information between two variables is defined as $\text{PMI}(\bm{x}_1,\bm{x}_2) \!=\! \tfrac{p(\bm{x}_1|\bm{x}_2)}{p(\bm{x}_1)}$. Each PMI term in Eq~\ref{eq:elbo_DO} captures the information that one sequence element $\bm{x}_{i\sm1}$ has regarding the next  $\bm{x}_{i}$, \textit{over and above} any information from earlier elements $\bm{x}_{<i\sm1}$ and the latent state $z$. Since the dropout-adapted ELBO (Eq~\ref{eq:elbo_DO}) is maximised w.r.t.\ $\theta$ and $\phi$, the weighted PMI term is minimised, which is only achievable by \textit{increasing the information learned by the model}: extracted from $\bm{x}_{<i\sm1}$ or captured in $z$.
Interestingly this shows that word dropout quite literally follows the earlier intuition and `weakens' the decoder by restricting \textit{information}. Thus, word dropout leads to a looser variational objective, i.e. $\text{ELBO}^{{\textbf{WD}}}_{\theta, \phi}(\bm{x}) \leq \text{ELBO}^{{\textbf{AR}}}_{\theta, \phi}(\bm{x})$. \citet{rainforth2018tighter} has argued that tighter bounds are not necessarily better. Hence, our analysis provides a justification for using word dropout.
\paragraph{The minimax objective:}

Under uniform dropout, all $d_i$ are equal over a sequence, and using dropout to `push information into $z$' is applied evenly. However, some sequence elements may hold more information than others, e.g.\ in language, given syntactic rules and a summary of the previous text, some words may be highly predictable without specifically knowing which word preceded them, whereas others may depend heavily on their predecessor despite the other information. This suggests that information content may be non-uniform and so applying dropout non-uniformly may be more appropriate. We, therefore, propose \textit{targeted} dropout of elements according to their `incremental' information content defined by the PMI terms, i.e.\ that not yet learned by the model. Further, since such incremental information content is not explicitly computed and will vary over training as the model learns, we train an adaptive dropout schedule to continually \text{maximise} the information dropped, and so minimize Eq~\ref{eq:elbo_DO}, with respect to $d_i$. Since Eq~\ref{eq:elbo_DO} is maximised with respect to all other parameters, this leads to a minimax objective.

We introduce the two `players' in the proposed minimax setting: $\text{VAE}_{\phi, \theta}$, trained to maximize the ELBO, and the adversary $\text{A}_{\psi}$ trained to do the opposite by determining each probability $d_i$ of dropping out the previous sequence element $\bm{x}_{i\sm1}$, preventing it from helping to predict its successor $\bm{x}_i$. The full objective for an observed sequence is given by
\begin{align} \label{eq-minimax}
     \max_{\phi, \theta} \; \min_{\psi} \; \mathbb{E}_{\bm{x} \sim p_{*}(\bm{x})} \left[\mathcal{L}_{\psi, \phi, \theta}(\bm{x})  + \mathcal{R}_{\psi}(\bm{x})\right]
\end{align}
where $\mathcal{R}_{\psi}(\bm{x})$ is a regularization term explained below (Section \ref{sec:optimisation}) and
\begin{align} \label{eq-modified-elbo}
    \mathcal{L}_{\psi, \phi, \theta} = 
    \mathbb{E}&_{q_{\phi}(z|\bm{x})}\left[\sum_i \log p_{\theta}(\bm{x}_i|\text{mask}_{K,\psi}(\bm{x}_{i\sm1};\bm{x}), \bm{x}_{<i\sm1}, z)\right] - \text{ KL}(q_{\phi}(z|\bm{x})||p_0(z))
\end{align}
$\mathcal{L}_{\psi, \phi, \theta}(\bm{x})$ is a modified version of the ELBO in Equation~\ref{eq:elbo_DO} with the additional parameters $\psi$ of the adversary $\text{A}_{\psi}$. The adversary manifests via the masking operator mask$_{K,\psi}(\cdot\,;\bm{x})$, conditioned on the entire input sequence $\bm{x}$. Since unconstrained minimisation of Eq~\ref{eq-modified-elbo} w.r.t $\psi$ would cause all elements to be dropped out, a constraint $K$ is introduced such that exactly $K$ elements of $\bm{x}$ are dropped during autoregressive decoding. In practice, $K$ is sequence length-dependent, treated as a hyperparameter that controls the level of adversarial regularization.
The adversary plays no role at test time.

\subsection{Implementation} 
\label{sec:method}

A high-level overview of our framework is shown in Figure~\ref{fig-adversarial-dropout}. 
An encoder maps the input sequence into a hidden representation to produce the parameters of $q_\phi(z|x)$, from which $z$ is sampled using the `reparametrization trick' \citep{kingma2013auto,rezende2014stochastic}. We implement the decoder using a unidirectional LSTM \citep{hochreiter1997long} for consistency with the RNN-VAE network of \cite{bowman2016}, because LSTMs are the most popular backbone for sequence VAEs and to facilitate comparison with prior work. Inspired by the recent work of \citep{dieng2019avoiding}, we choose a \textit{Double-LSTM} recurrent unit
that elucidates the \emph{skip-connections} to promote higher latent information content flow (see Appendix \ref{sec:app_double-lstm} for details). 

As described in Section \ref{sec:awd_theory}, the main innovation of our proposed method is the \textit{adversarial} training procedure: instead of dropping out decoder inputs uniformly at random, we introduce a trainable {adversary} to drop words most important for the VAE to reconstruct the original sequence.
Below we describe implementation details of the adversary, specifying the relevant components in  Figure~\ref{fig-adversarial-dropout}. A more comprehensive figure specific to the implemented architecture is provided in Appendix \ref{sec:app_architecture}. Note that the theoretical framework in Section \ref{sec:awd_theory} permits many alternative approaches for generating per word dropout probabilities $d_i$, such as attention mechanisms mentioned above. We leave further refinement of the adversarial architecture to future work. 

\paragraph{Producing dropout scores}
To determine which elements to drop out of a sequence, the adversary $\text{A}_{\psi}$ samples a \textit{score} $s_i \!\in\! \mathbb{R}$ for each sequence element $\bm{x}_i$. Sequence elements with the $K$ smallest scores are dropped. 
The scores are sampled from distributions $p_{\psi}(s_i|\bm{x})$ conditioned on the input sequence $\bm{x}$, modelled as a series of Gaussians with mean ($\mu$) and variance ($\sigma$) parameterised by the outputs of a unidirectional LSTM. 
Scores are generated using the reparameterization trick, reducing gradient variance \citep{kingma2013auto}. 
Thus, $p_{\psi}(s_i|\bm{x})$ is described as
\begin{align*} 
    s_i|\bm{x} \sim \mathcal{N}(\mu_i, \sigma_i ; \bm{x}, \psi) \quad  \qquad\qquad 
    \text{where,} \quad [\mu_i, \sigma_i] = \text{Linear}_{\psi}(\text{LSTM}_{i,\psi}(\bm{x}))
\end{align*}
The stochastic generation of scores gives the adversary an `exploratory' capability during training, e.g.\ preventing it becoming `stuck' on a set of sequence elements with high information content that the model cannot learn, meaning that their PMI terms (in Eq~\ref{eq:elbo_DO}) remain large and the adversary repeatedly selects them to minimise the ELBO.

\paragraph{Top-K word selection} Based on the sampled scores $\bm{s} \!=\! \{s_i\}_{i=1}^T$, the subset of $K$ words with the smallest scores are masked during decoding. To estimate gradients of the objective with respect to the parameters $\psi$, we use a stochastic softmax trick from \citet{paulus2020gradient}: during training, the stochastic subset selected in the forward pass is relaxed to admit a (biased) reparameterization gradient in the backward pass. 
In our experiments, we use a straight-through variant \citep{bengio2013estimating, paulus2020rao} of the trick for our method to resemble word dropout, i.e.\ to produce discrete values $\in\! \{0,1\}$. The relaxation is only used in the backward pass to compute the gradient estimator.

\paragraph{Gradient reversal} The final component of the adversarial network is the \emph{gradient reversal} layer \citep{ganin2016domain}.
In the forward pass, the layer performs no transformation to the input; in the backward pass, the gradients are negated.
Gradient reversal offers a computationally simple method to ensure that the parameters $\psi$ of $\text{A}_{\psi}$ are updated such that ELBO is minimized. If $\bm{I}$ is an identity matrix, the `pseudo-function' of gradient reversal can be described as
\begin{align} 
    f(\bm{x}) = \bm{x}                                         \qquad\text{(forward pass)}
    \qquad\qquad\qquad
    \frac{\partial f(\bm{x})}{\partial \bm{x}} = -\bm{I}       \qquad\text{(backward pass)}
    \nonumber
\end{align}

\begin{table*}[t!]
    \centering
    {\small
    \setlength{\tabcolsep}{4pt}
    \begin{tabular}{lcccc|cccc}
        \toprule
          & PPL $\downarrow$ & -ELBO $\downarrow$ & KL $\uparrow$ & MI $\uparrow$ & PPL $\downarrow$ & -ELBO $\downarrow$ & KL $\uparrow$ & MI $\uparrow$\\
        \midrule
        \textbf{Existing sequence VAEs} & \multicolumn{4}{c}{\textbf{Yahoo}} & \multicolumn{4}{c}{\textbf{Yelp}} \\
         CNN \citep{yang2017improved}            & 63.90 & - & 10.0 & -     & 41.1 & - & \textbf{7.6} & - \\ 
         Lagging \citep{he2018lagging}           & - & \textbf{328.4} (0.2) & 5.7 (0.7) & 2.90   & - & 357.2 (0.1) & 3.8 (0.2) & 2.4 \\ 
         SA \citep{kim2018semi}                  & 60.40 & - & 7.19 & -       & - & - & - & -\\ 
         Skip \citep{dieng2019avoiding}                  & 60.90 & 330.3 & \textbf{15.05} & 7.47       & - & - & - & -\\ 
         FBP \citep{li2019surprisingly}           & 59.51 & 330.3 & 15.02 & -  & - & - & - & - \\
         \midrule
         \textbf{Our sequence VAE} &  \\
         unregularized                           & 60.30 & 328.8 (0.2) & 4.2 (0.2)  & 3.14      & 40.1 & 356.4 (0.2) & 2.3 (0.2) & 1.0 \\
         + word dropout [0.4]                 & 59.55  & 329.5 (0.4)  & 14.4 (0.4) & \textbf{13.6}       & 38.5 & \textbf{354.2} (0.3) & 5.9 (0.4) & 4.9\\
         + AWD (\emph{ours}) [0.3] & \textbf{59.05}  & \textbf{328.4} (0.3) & 14.4 (0.4) & \textbf{13.6}    & \textbf{38.2} & \textbf{354.2} (0.3) & \textbf{6.5} (0.4) & \textbf{5.8} \\
        \bottomrule
    \end{tabular}
    }
    \caption{\textbf{Results of text modeling on the Yahoo and Yelp datasets.} Standard deviations are provided in the brackets.
    Squared bracket contains the dropout rate $DR$.
    PPL -- perplexity; ELBO -- evidence lower bound; KL - in Eq~(\ref{eq-modified-elbo});  MI -- mutual information $I(\bm{X};Z)$ from Section \ref{sec:bg}.}
    \label{tab-lm-results}

\end{table*}%

\subsection{Optimization challenges}
\label{sec:optimisation}

Since our adversarial dropout network is fully differentiable, it is readily optimized by gradient methods, such as backpropagation. However, the magnitudes of the word dropout scores $\bm{s}$ sampled from distributions parameterized by an LSTM are unconstrained. 
Controlling the magnitude of the scores was found to be important for the adversary to maintain its exploratory capability (discussed above). We therefore add a KL-divergence between the distribution of each score $p(s_i|\bm{x}_i)$ and a standard Gaussian $p_0(s_i)\!=\!\mathcal{N}(s_i;0,1)$ to `regularize' scores in Eq~(\ref{eq-minimax}), subject to a scalar $\lambda \!>\! 0$:

\begin{align} 
    \label{eq-reg-term}
    \mathcal{R}_{\psi}(\bm{x}) = \lambda \sum_i \text{ KL}(p_{\psi}(s_i|\bm{x})\,||\,p_0(s_i))
\end{align}
\vspace{-0.4cm}
\paragraph{Adversarial vs.\ random word dropout}
Standard word dropout \citep{iyyer2015deep,bowman2016} can be seen as a special case of adversarial word dropout.
By setting $\lambda$ sufficiently high, the regularization term in Eq~\ref{eq-reg-term} can be made to dominate such that $p_{\psi}(s_i|\bm{x}) \approx p_0(s_i)$ and all scores $s_i$ are effectively sampled from a standard Gaussian $p_0(\bm{s}_i)$ and words are dropped approximately uniformly.
On the other hand, for small values of $\lambda$, $\text{A}_{\psi}$ will learn which words $\text{VAE}_{\phi, \theta}$ relies upon most to accurately decode the sequence and target those for drop out. 
As a result, whilst adding another hyperparameter to calibrate, $\lambda$ offers a simple way to moderate the difference between adversarial word dropout and standard uniform word dropout,
which we consider in our experiments.

\begin{figure}[b] 
	\centering	
	\includegraphics[width=0.95\textwidth]{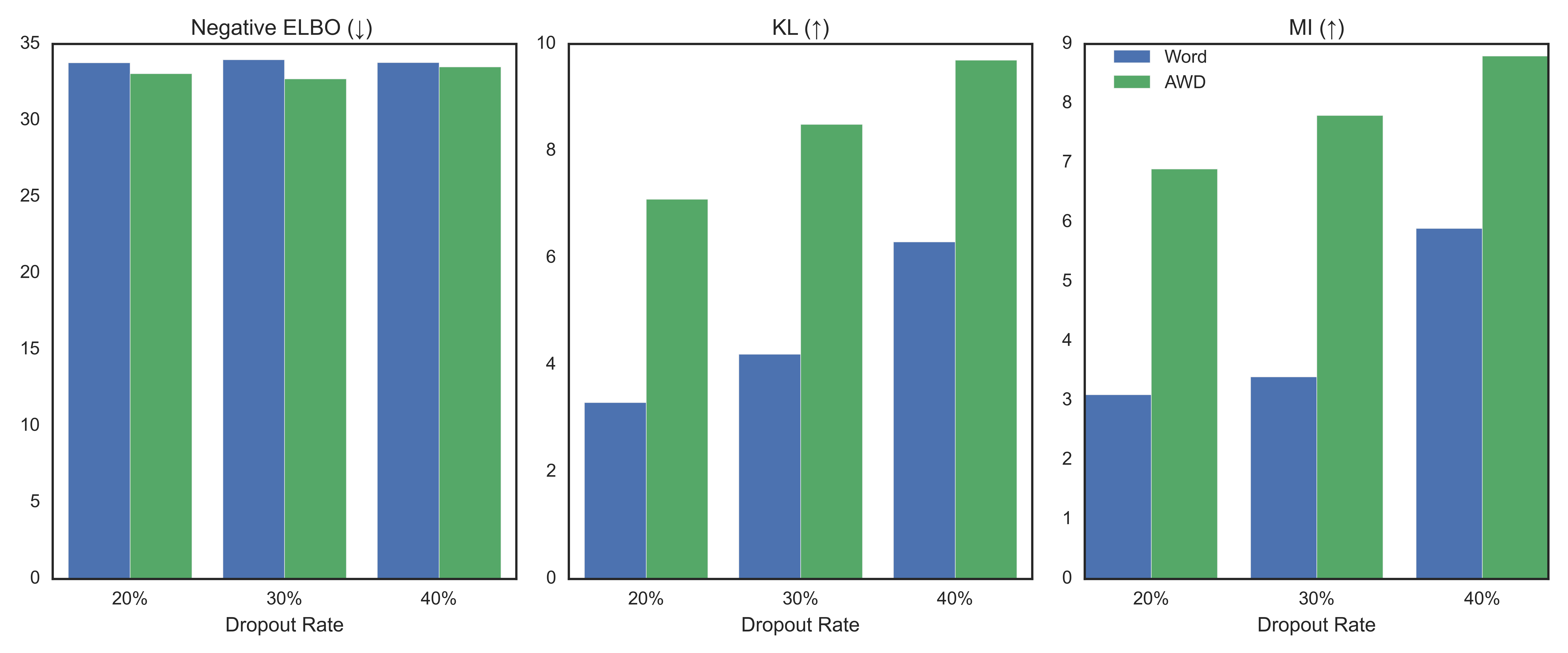}
    \caption{Adversarial vs uniform word dropout (SNLI), with ablation of the various dropout rates. }
	\label{fig-ablation-studies}
	\vspace{-10pt}
\end{figure}

\section{Experiments} \label{sec:exp}

Here we present the results of our experiments that: \textbf{(i)} demonstrate that a sequence VAE trained with adversarial word dropout (AWD) outperforms other sequence VAEs; it  achieves  improved sentence modeling performance and/or improved informativeness of the latent space; \textbf{(ii)} examine the contributions and behaviour of the adversarial network's components and hyperparameters, and \textbf{(iii)} qualitatively study the trained adversary and VAE.

\paragraph{Datasets}
We conducted experiments on 4 different datasets: \emph{Yahoo} questions and answers \citep{yang2017improved}, \emph{Yelp} reviews \citep{yang2017improved}, Penn Tree Bank (PTB) \cite{marcus-etal-1993-building} and downsampled Stanford Natural Language
Inference (\emph{SNLI}) corpus \citep{bowman-etal-2015-large, li2019surprisingly}.
Yahoo, Yelp and PTB datasets were used in many previous works \citep{yang2017improved,kim2018semi,he2018lagging,fu-etal-2019-cyclical,dieng2019avoiding} hence were used to benchmark our proposed method against comparable related works and also against standard word dropout. Yahoo, Yelp and PTB contain sentences with average lengths of 78, 96 and 22 words respectively, while SNLI sentences are much shorter with an average length of 9 words and are more suitable for qualitative studies. Yahoo, Yelp and SNLI datasets contain 100K sentences in the training set, 10K in the validation set, and
10K in the test set, while PTB is much smaller with a total of 42K sentences.

\paragraph{Experimental setup} We (re-)implement the standard VAE, a VAE with standard uniform word dropout and a VAE with our adversarial dropout method. On each dataset, we performed the same grid search over both learning rate (from $\{0.0001, 0.001, 0.1, 1\}$) and dropout rate R (from $\{0.2, 0.3, 0.4, 0.5\}$) for both the word dropout baseline and our method. This gives 16 different hyper-parameter configurations for each method on each dataset. For training, we also use an exponential learning decay of 0.96 as in \citep{li2019exponential},  increased the hidden state sze of the decoder LSTM from 1024 to 2048 (except on SNLI), applied Polyak averaging \citep{polyak1992acceleration} with a coefficient of 0.9995 and used \textit{KL annealing} \citep{bowman2016}. 
 
We apply early stopping based on validation ELBO and repeat each experiment for five different random seeds to report standard deviations. All experiments are performed on a 12GB Nvidia TitanXP GPU with an average run time of 4 hours for Yelp and Yahoo and 1 hour for SNLI. 

\paragraph{Metrics} Overall model performance is assessed based on two main aspects: \textbf{(i)} \emph{sentence modeling} -- measured with respect to ELBO and perplexity (PPL). ELBO and PPL quantify the ability of the trained decoder to recognise or generate natural language, reflecting the quality of density estimation; \textbf{(ii)} \emph{latent space informativeness} -- measured with respect to the KL term in Eq~(\ref{eq-modified-elbo}) and mutual information $I(\bm{X};Z)$ between observed and latent variables. $I(\bm{X};Z)$ is computed using the procedure of \cite{hoffman2016elbo}, as implemented by \cite{dieng2019avoiding}.
As in comparable works \citep{yang2017improved,kim2018semi,he2018lagging,dieng2019avoiding,li2019surprisingly}, hyperparameters are set based on a balanced assessment of these metrics. 

\paragraph{Adversary hyperparameters} To apply the global dropout rate $R$ to the number $K$ of elements to be dropped out we use $K = \emph{round}(R \times T)$ where \emph{round} computes the closest integer. The additional hyperparameter $\lambda$ in Eq~\ref{eq-reg-term} was set globally for all datasets to $\lambda = 1$. This was based on an initial exploratory analysis on the Yahoo dataset, where we grid searched $\lambda$ in $\{0.001, 0.01, 0.1, 1, 5, 10\}$ for various combinations of dropout and learning rates to find that $\lambda=1$ consistently achieved best validation performance. Please refer to the detailed analysis of $\lambda$ in the appendix.

\subsection{Quantitative analysis} 
Table \ref{tab-lm-results} compares 
\textbf{(i)} the VAE trained with no dropout;
\textbf{(ii)} the VAE trained with uniform word dropout \citep{bowman2016} (dropout rate $R \!=\! 0.4$ found to be best); 
\textbf{(iii)} a VAE trained with adversarial word dropout ($R \!=\! 0.3$ found best); and
\textbf{(iv)} previously reported results for comparable models (Section \ref{sec:related-work}). Our adversarial dropout method trains models that achieve better sentence modeling (lower ELBO, PPL) with an equally informative latent space (Yahoo, PTB) or more informative latent space (higher KL, MI) while maintaining sentence modeling performance (Yelp) and improves both metrics on SNLI (Appendix Table \ref{SNLI comparison}) . Thus, it allows users to more effectively trade-off sentence modelling and informativenes of the latent space than standard word dropout. The gains are modest in size, but larger on SNLI and PTB, and comparable to those improvements reported in previous word, e.g., \citep{he2018lagging} (Appendix Table \ref{PTB comparison}, \ref{SNLI comparison}).
Our method also compares favourably to previous models, consistently achieving an improved balance between sentence modeling performance (e.g.\ PPL)  and  an informative latent space (e.g.\  MI).
We note also that the vanilla VAE does not obtain a lower perplexity than either VAE with word dropout, as might be anticipated if dropout were an arbitrary `regularisation' method that may improve the latent space but at a cost to sequence modeling performance. This supports our theoretical analysis indicating that word dropout is not a typical `regularizer' (Section \ref{sec:awd_theory}), rather that it disrupts the flow of available information during training of the autoregressive decoder, forcing it to compensate by storing information in the latent space. 
Figure~\ref{fig-ablation-studies} compares adversarial and uniform word dropout, varying the dropout rate on the SNLI dataset. For any given dropout rate, adversarial dropout learns a more informative latent space (with higher KL and MI) metrics with comparable to lower negative ELBO. 

\begin{figure}[h!] 
	\centering	
	\includegraphics[width=0.7\textwidth]{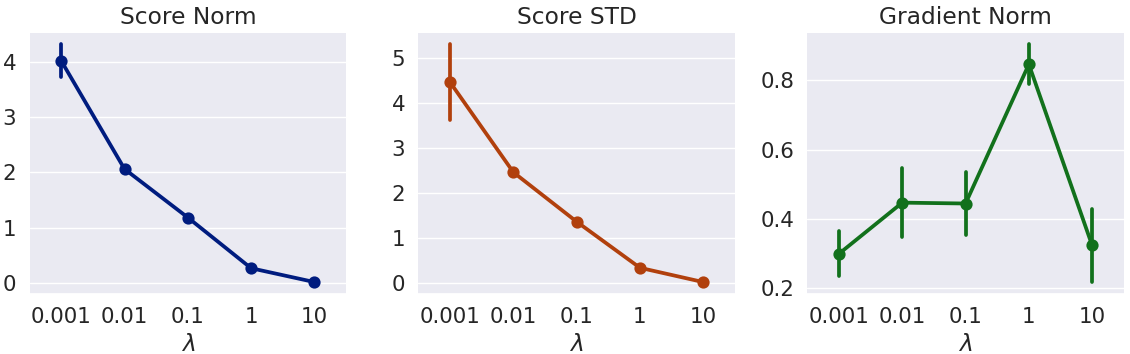}
    \caption{The role of $\lambda$. On Yahoo dataset, we computed various statistics from the data collected across iterations during one training run; \emph{(left and mid)} Mean $\ell_1$-norm and standard deviation of dropout score vectors ($\bm{s}$); \emph{(right)} Gradient norm of the adversary -- signifies the magnitude of the parameter updates and hence the quality of learning.
    }
	\label{fig-lambda-studies}
\end{figure}

\vspace{-0.2cm}
\paragraph{The role of $\lambda$} The hyperparameter $\lambda$ can be seen to determine the \emph{exploration-exploitation trade-off} of the adversary. As shown in Figure~\ref{fig-lambda-studies}, for small values of $\lambda$, the magnitudes and standard deviations of word dropout scores grow large, causing the distribution of dropped words to become concentrated.
For large values of $\lambda$, scores become very small with low variance and the adversary converges towards uniform word dropout.
The right part of Figure~\ref{fig-lambda-studies} depicts why $\lambda\!=\!1$ is a good choice.

\begin{figure}[h!]
     \centering
     \begin{subfigure}[b]{0.48\textwidth}
         \centering
         \includegraphics[width=\textwidth]{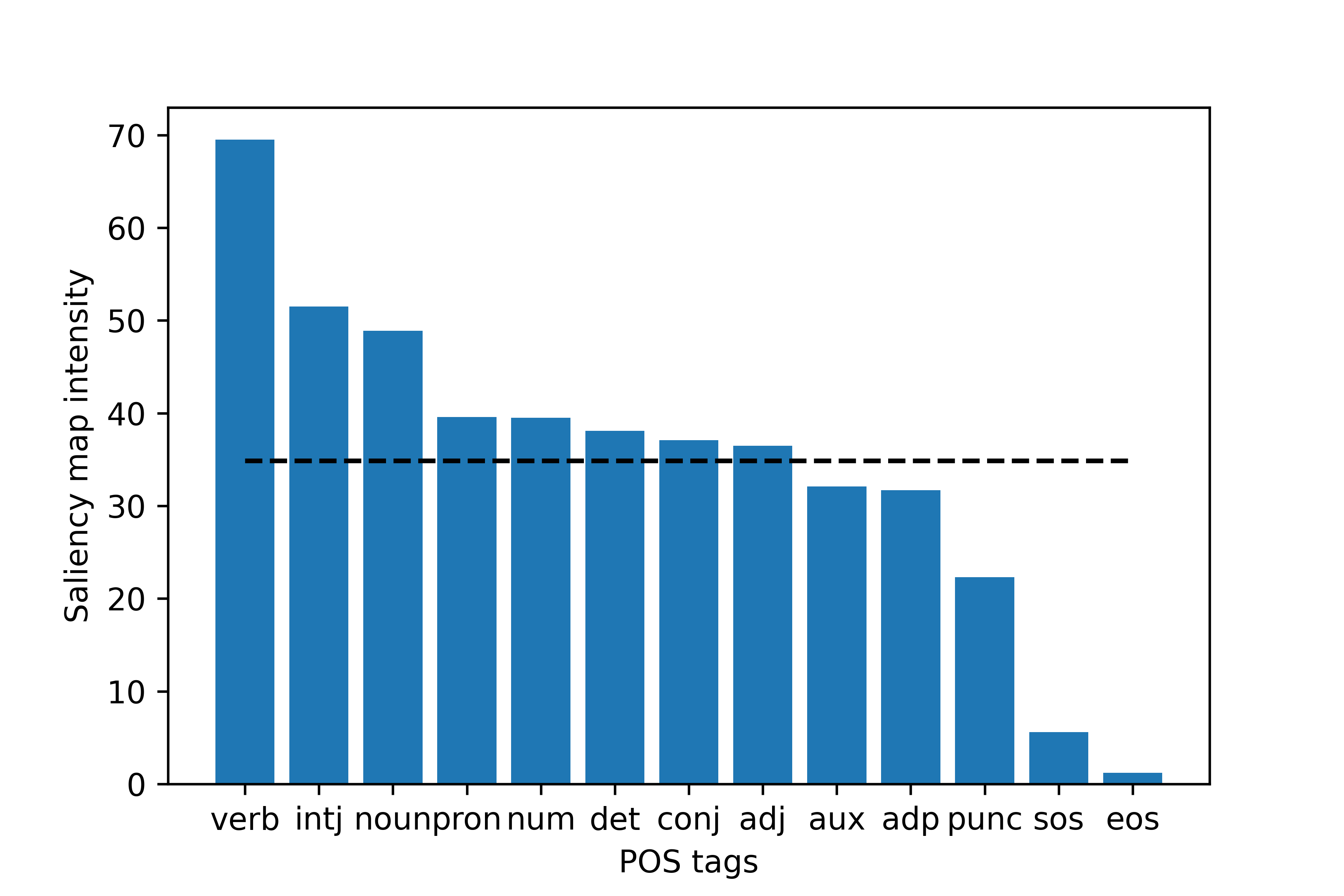}
         \caption{Quantitative analysis of dropout saliency map across different part-of-speech (POS) tags. Dotted horizontal line represents the average saliency across all words. }
         \label{bert-f1}
     \end{subfigure}
     \hfill
     \begin{subfigure}[b]{0.48\textwidth}
         \centering
         \includegraphics[width=\textwidth]{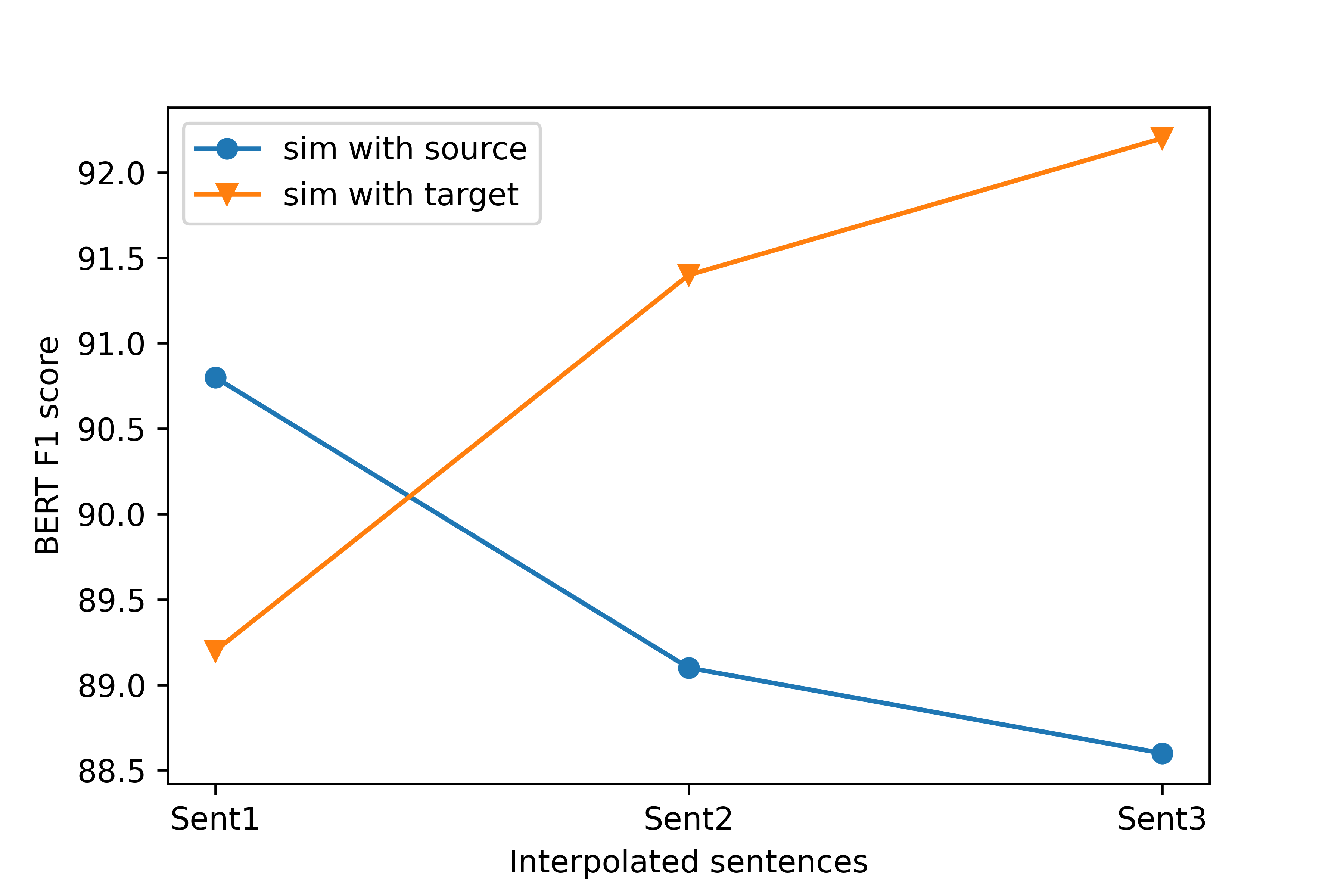}
         \caption{BERT $F_1$ score computed between the interpolated sentences with the source and target sentence. \textit{Sent} represents the interpolated sentences (set to three in our analysis).}
         \label{fig:sal}
     \end{subfigure}
     \hfill
        \caption{Quantitative analysis of the interpolations and saliency map presented in the paper.}
        \label{fig:bert-n-sal}
\end{figure}

\paragraph{Qualitative analysis}  We obtain further insight into what the adversary learns by analyzing the word dropout scores for different sentences. Table \ref{tab-drop-outs} shows that the adversary applies lower dropout probabilities to less informative words such as `unknown' tokens that replace all out-of-dictionary words, and so offer little information about the next word. Depending on the data semantics, the adversary selects different types of words: for SNLI, verbs tend to be picked that explain the activity, e.g.\ \emph{working} and \emph{participating}; for Yahoo, words are identified that carry question semantics, e.g.\ \emph{what}, \emph{how}, \emph{if}, and \emph{when}. Figure \ref{fig:bert-n-sal} (a) shows a quantitative analysis of dropout saliency map across different part-of-speech (POS) tags. Verb (verb), interjections (intj), and nouns (noun) have higher saliency scores (higher chances of being dropped) compared to punctuation (punc), determiners (det), and the start (sos) and end tokens (eos) which are relatively easier to predict given previous words. We also show that adversarial training learns a useful generative model with meaningful latent space by interpolating between sentences (Table \ref{tab-interpolation}). Computing BERT $F_1$ score \cite{zhang2019bertscore} between the interpolated sentences with the source and target sentence shows the increasing trend toward the target sentence for each interpolation (as each interpolated sentence is getting away from the source and closer to the target) and decreasing for the source sentence (Figure \ref{fig:bert-n-sal} (b)).

\begin{table}[t!]
    \centering
    {\renewcommand{\arraystretch}{1.5}
    \begin{tabular}{@{}l@{}}
    \toprule
    \begin{CJK*}{UTF8}{gbsn}
    {\setlength{\fboxsep}{0.34pt}\colorbox{white!0}{\parbox{0.99\textwidth}{
    \colorbox{red!0.0}{\strut <sos>} \colorbox{red!13.618934194312317}{\strut the} \fbox{\colorbox{red!69.21610067395929}{\strut person}} \fbox{\colorbox{red!100.0}{\strut giving}} \colorbox{red!64.29599135349474}{\strut the} \colorbox{red!62.63966240078204}{\strut demonstration} \colorbox{red!5.960305932080871}{\strut is} \colorbox{red!41.211336835076175}{\strut a} \colorbox{red!57.44349825486538}{\strut man} \colorbox{red!4.771411459373127}{\strut <eos>}
    }}}
    \end{CJK*} \\
    \begin{CJK*}{UTF8}{gbsn}
    {\setlength{\fboxsep}{0.3pt}\colorbox{white!0}{\parbox{0.99\textwidth}{
    \colorbox{red!2.792607802874741}{\strut <sos>} \colorbox{red!0.0}{\strut a} \colorbox{red!40.664603638256644}{\strut dog} \fbox{\colorbox{red!100.0}{\strut participating}} \colorbox{red!29.627787257425453}{\strut in} \colorbox{red!27.762385400699884}{\strut a} \colorbox{red!36.98527923185932}{\strut race} \colorbox{red!1.3384619833993696}{\strut while} \fbox{\colorbox{red!87.96251843711138}{\strut wearing}} \colorbox{red!71.1553923128091}{\strut the} \colorbox{red!56.417271596726145}{\strut number} \colorbox{red!30.62613876276137}{\strut 6} \colorbox{red!3.572895277207394}{\strut .} \colorbox{red!6.915579720623526}{\strut <eos>}
    }}}
    \end{CJK*} \\
    \begin{CJK*}{UTF8}{gbsn}
    {\setlength{\fboxsep}{0.34pt}\colorbox{white!0}{\parbox{0.99\textwidth}{
    \colorbox{red!26.78751436908658}{\strut <sos>} \colorbox{red!47.498452559908046}{\strut three} \colorbox{red!60.91903203937867}{\strut women} \colorbox{red!29.1207592772718}{\strut are} \fbox{\colorbox{red!100.0}{\strut talking}} \colorbox{red!24.007427712441427}{\strut about} \fbox{\colorbox{red!68.6668435169629}{\strut their}} \colorbox{red!32.821646476257854}{\strut matching} \colorbox{red!17.1721637633743}{\strut wigs} \colorbox{red!49.57644354054294}{\strut and} \fbox{\colorbox{red!64.7513779585581}{\strut silver}} \colorbox{red!41.08409231585463}{\strut leggings} \colorbox{red!0.0}{\strut .} \colorbox{red!29.629498629410207}{\strut <eos>}
    }}}
    \end{CJK*} \\
    \midrule
    \begin{CJK*}{UTF8}{gbsn}
    {\setlength{\fboxsep}{0.3pt}\colorbox{white!0}{\parbox{0.99\textwidth}{
    \colorbox{red!0.0}{\strut <sos>} \colorbox{red!19.523892066793792}{\strut is} \fbox{\colorbox{red!45.4671520194141}{\strut there}} \colorbox{red!25.70636158779685}{\strut any} \fbox{\colorbox{red!55.05286878141792}{\strut truth}} \colorbox{red!29.883862021147507}{\strut to} \colorbox{red!33.74934997399896}{\strut the} \colorbox{red!26.00681805049979}{\strut rumor} \colorbox{red!22.99647541457213}{\strut that} \colorbox{red!39.72381117466921}{\strut the} \colorbox{red!14.410354191945459}{\strut host} \colorbox{red!24.446755648003702}{\strut of} \colorbox{red!22.135552088750217}{\strut the} \colorbox{red!4.3566187091928015}{\strut \_unk} \fbox{\colorbox{red!59.22459120587046}{\strut on}} \colorbox{red!11.983590454729296}{\strut \_unk} \colorbox{red!21.442191020974178}{\strut ,} \colorbox{red!33.9284682498411}{\strut greg} \colorbox{red!8.511007106950942}{\strut \_unk} \colorbox{red!31.62304269948575}{\strut is} \colorbox{red!33.708903911712014}{\strut dead} \colorbox{red!25.076558617900275}{\strut ?} \colorbox{red!13.191194314439244}{\strut \_unk} \fbox{\colorbox{red!76.11371121511527}{\strut has}} \fbox{\colorbox{red!70.79216501993413}{\strut been}} \colorbox{red!34.795169584561165}{\strut down} \colorbox{red!33.483561564684805}{\strut and} \fbox{\colorbox{red!50.488241751892296}{\strut there}} \colorbox{red!26.80996128734037}{\strut is} \colorbox{red!32.74975443462182}{\strut no} \fbox{\colorbox{red!54.27861558906801}{\strut new}} \colorbox{red!17.131796382966428}{\strut information} \colorbox{red!39.37713064078119}{\strut on} \colorbox{red!23.2969318772751}{\strut the} \colorbox{red!33.876466169757904}{\strut web} \fbox{\colorbox{red!41.09897729242503}{\strut about}} \colorbox{red!23.152481654821756}{\strut the} \colorbox{red!10.920436817472705}{\strut future} \colorbox{red!39.51580285433639}{\strut of} \colorbox{red!26.428612700063553}{\strut the} \colorbox{red!13.555208875021663}{\strut site} \colorbox{red!36.54012827179754}{\strut or} \colorbox{red!28.803374357196503}{\strut the} \colorbox{red!6.338475761252667}{\strut \_unk} \colorbox{red!18.36251227826891}{\strut ,} \colorbox{red!31.73860287744843}{\strut greg} \colorbox{red!9.054139943375517}{\strut \_unk} \colorbox{red!33.66267984052695}{\strut .} \fbox{\colorbox{red!85.13896111400013}{\strut what}} \colorbox{red!35.5752007858092}{\strut 's} \fbox{\colorbox{red!92.72548679724967}{\strut going}} \colorbox{red!17.160686427457094}{\strut on} \colorbox{red!27.555324435199623}{\strut ?} \fbox{\colorbox{red!100.0}{\strut how}} \fbox{\colorbox{red!71.76864852371872}{\strut could}} \colorbox{red!5.41977234644942}{\strut \_unk} \colorbox{red!30.895013578320913}{\strut not} \colorbox{red!13.971225515687294}{\strut comment} \colorbox{red!28.318021609753274}{\strut on} \colorbox{red!20.015022823135155}{\strut the} \colorbox{red!8.990581845496038}{\strut past} \colorbox{red!17.64026116600219}{\strut us} \colorbox{red!17.17224244525336}{\strut election} \colorbox{red!19.483446004506845}{\strut (} \fbox{\colorbox{red!53.296354076385285}{\strut nov.}} \colorbox{red!10.400416016640662}{\strut 7th} \colorbox{red!18.668746749869996}{\strut )} \colorbox{red!26.053042121684882}{\strut nor} \colorbox{red!29.652741665222166}{\strut the} \colorbox{red!27.23175593690415}{\strut iranian} \colorbox{red!29.85497197665684}{\strut nuclear} \colorbox{red!10.007511411567577}{\strut program} \colorbox{red!19.49500202230311}{\strut ?} \fbox{\colorbox{red!59.49615762408274}{\strut something}} \fbox{\colorbox{red!70.34725833477785}{\strut must}} \colorbox{red!25.59657941873232}{\strut be} \colorbox{red!30.536777026636614}{\strut wrong} \colorbox{red!17.27046859652164}{\strut ,} \fbox{\colorbox{red!44.143987981741496}{\strut where}} \colorbox{red!38.04241058531231}{\strut is} \colorbox{red!12.497833246663191}{\strut this} \colorbox{red!17.732709308372335}{\strut guy} \colorbox{red!20.50037557057837}{\strut ?} \fbox{\colorbox{red!62.73184260703762}{\strut internet}} \colorbox{red!33.07332293291732}{\strut chat} \colorbox{red!16.819783902467194}{\strut rooms} \fbox{\colorbox{red!52.533656901831634}{\strut have}} \colorbox{red!38.78199572427342}{\strut noted} \colorbox{red!9.192812156930714}{\strut that} \colorbox{red!7.332293291731673}{\strut \_unk} \colorbox{red!21.57508522563124}{\strut was} \colorbox{red!12.40538510429306}{\strut injured} \colorbox{red!31.85994106430924}{\strut in} \colorbox{red!27.353094123764947}{\strut a} \colorbox{red!9.487490610735534}{\strut fly} \colorbox{red!11.515571733980465}{\strut fishing} \colorbox{red!20.52926561506905}{\strut accident} \colorbox{red!18.125613913445434}{\strut earlier} \fbox{\colorbox{red!45.83694458889467}{\strut this}} \colorbox{red!12.838735771653091}{\strut month} \colorbox{red!18.818974981221487}{\strut .} \fbox{\colorbox{red!49.62154041717225}{\strut details}} \colorbox{red!25.31345698272375}{\strut are} \colorbox{red!22.216444213324095}{\strut unknown} \colorbox{red!26.798405269544105}{\strut .} \fbox{\colorbox{red!95.30825677471543}{\strut can}} \fbox{\colorbox{red!83.37666840006933}{\strut anyone}} \colorbox{red!37.87484832726642}{\strut clear} \colorbox{red!20.21725313456983}{\strut this} \colorbox{red!11.001328942046555}{\strut up} \colorbox{red!23.73606055353325}{\strut ?} \colorbox{red!0.9302594325995273}{\strut <eos>}
    }}}
    \end{CJK*} \\
    \begin{CJK*}{UTF8}{gbsn}
     {\setlength{\fboxsep}{0.3pt}\colorbox{white!0}{\parbox{0.99\textwidth}{
    \colorbox{red!0.33098882912702265}{\strut <sos>} \colorbox{red!28.291270169631773}{\strut is} \colorbox{red!52.25486139842779}{\strut it} \colorbox{red!54.861398427803046}{\strut okay} \fbox{\colorbox{red!87.91890773686387}{\strut if}} \colorbox{red!37.89822093504343}{\strut i} \colorbox{red!36.18535374431112}{\strut mind} \colorbox{red!9.929664873810495}{\strut \_unk} \fbox{\colorbox{red!64.82416218452627}{\strut with}} \colorbox{red!50.15308233347124}{\strut my} \colorbox{red!40.22341745966072}{\strut goldfish} \colorbox{red!34.59660736450144}{\strut ?} \fbox{\colorbox{red!69.81381878361606}{\strut yes}} \colorbox{red!25.75093090608192}{\strut ,} \colorbox{red!48.2829954489036}{\strut but} \fbox{\colorbox{red!100.0}{\strut be}} \colorbox{red!25.891601158460887}{\strut careful} \colorbox{red!23.475382705833667}{\strut ,} \colorbox{red!49.424906909391794}{\strut you} \colorbox{red!56.06950765411667}{\strut may} \colorbox{red!59.99999999999999}{\strut be} \colorbox{red!49.3256102606537}{\strut overwhelmed} \fbox{\colorbox{red!63.94704178733967}{\strut when}} \colorbox{red!30.22755482002482}{\strut you} \colorbox{red!55.51510136532891}{\strut connect} \colorbox{red!30.128258171286717}{\strut to} \fbox{\colorbox{red!79.16425320645428}{\strut such}} \colorbox{red!41.348779478692585}{\strut a} \colorbox{red!21.985932974762093}{\strut clearly} \colorbox{red!27.9437318990484}{\strut superior} \colorbox{red!20.31443938767066}{\strut intellect} \colorbox{red!19.79313198179561}{\strut ...} \colorbox{red!0.0}{\strut <eos>}}}}
    \end{CJK*} \\
    \bottomrule
    \end{tabular}}
    \vspace{0.2cm}
    \caption{\text{Analysis of the Adversary.} For selected SNLI \emph{(upper)} and Yahoo \emph{(lower)} sentences, word dropout scores are from a trained adversary and normalized per sentence. Darker colouring indicates a higher dropout probability. Boxed words are those selected to be dropped.\\ <sos> = start of the sequence token; <eos> = end of the sequence token; \_unk = unknown token.}
    \label{tab-drop-outs}
    \vspace{-0.0cm}
\end{table}

\section{Conclusions} \label{sec:conclusio}
\vspace{-0.25cm}
In this work, we address the phenomenon of \textit{posterior collapse} that occurs in autoregressive variational autoencoders (VAEs) used for sequential data.
To mitigate posterior collapse, we propose a novel adversarial approach that \textit{learns to drop words based on their information content}, leading to an improvement in both the learned latent structure and sequence modeling performance.
We theoretically derive the effect of word dropout to show that, during training, it removes \textit{incremental} information that each word provides about the next above that available from earlier words or the latent variable $z$. The only way for the model to compensate and minimise the incremental information lost is to \textit{learn more information in the latent space}. We believe this finding provides novel and interesting insight as a means of \textit{manipulating information} within a hierarchical latent variable model since dropout is used to `push' specific information into the latent variable. For future work, it may be interesting to extend our method to transformer-based architectures \citep{vaswani2017attention} that can exacerbate posterior collapse when used for decoding and self-supervised learning, where input (words in NLP, image patches in computer vision) are often masked randomly \citep{devlin2018bert, he2022masked}, but a learnt policy might improve performance or convergence. 

Sequence VAEs are a promising framework that, beyond purely autoregressive models \citep{brown2020language}, hold the prospect of controlled sequence generation. Improvements to such general methods may inevitably be used for good or ill, from the generation of targeted fake news on the one hand to the possibility of personalised human-computer interactions in, say, the medical domain on the other (e.g. for modeling sleep \citep{miladinovic2019spindle,nowak2021rapid}).
In future work, we plan to explore alternative implementation options, particularly of the adversary, and extend adversarial dropout to other domains, such as images, speech, or dynamical systems \citep{bauer2017efficient}.
\newline

\begin{table}[h!]
    \centering
    {\def\arraystretch{1}\setlength{\tabcolsep}{3pt}
    \begin{tabular}{@{}l@{}}
    \toprule
    \textbf{I 'm not sure what all the hype is about . i 've been here a few times and it 's just ok . nothing} \\
    \textbf{special . I would n't go out of my way to come here .}\vspace{0.1cm}\\
    \emph{\quad I 've been here a few times and it 's always been good . the food is good , but the service is} \\
    \emph{\quad not so great .}\\
    \emph{\quad I 've been here a few times and it 's always been good . i 've had the chicken and waffles and} \\
    \emph{\quad the service was good .}\\
    \emph{\quad Great place to go for a quick bite to eat . the food is great and the service is great . i have} \\
    \emph{\quad been here a few times and have never been disappointed .}\vspace{0.1cm}\\
    \textbf{Great food , great service , and great service . i 've been} \textbf{here a few times and have never been} \\
    \textbf{disappointed}\\
    \bottomrule
    \end{tabular}}
    \vspace{0.2cm}
    \caption{\text{Sentence interpolation (Yelp dataset).} Representations of two sentences \emph{(top, bottom)} are obtained by feeding them through an adversarially trained VAE encoder. Three linearly interpolated representations are passed to the VAE decoder and sentences generated by greedy sampling \emph{(middle)}.}
    \label{tab-interpolation}
\end{table}

\section{Acknowledgments}

We thank Taylor Berg-Kirkpatrick for his thoughtful insights and valuable feedback. 
Carl is gratefully supported by an ETH AI Centre Postdoctoral Fellowship, a responsible AI grant by the Haslerstiftung; Swiss National Science Foundation (project \# 201009), and an ETH Grant (ETH-19 21-1).


\newpage
\bibliographystyle{plainnat}
\bibliography{neurips_2022}

\section*{Checklist}

\begin{enumerate}

\item For all authors...
\begin{enumerate}
  \item Do the main claims made in the abstract and introduction accurately reflect the paper's contributions and scope?
    \answerYes{} (Section \ref{sec:method})
  \item Did you describe the limitations of your work?
    \answerYes{} (Section \ref{sec:conclusio})
  \item Did you discuss any potential negative societal impacts of your work?
    \answerYes{} (Section \ref{sec:conclusio})
  \item Have you read the ethics review guidelines and ensured that your paper conforms to them?
    \answerYes{} (All Sections)
\end{enumerate}

\item If you are including theoretical results...
\begin{enumerate}
  \item Did you state the full set of assumptions of all theoretical results?
    \answerYes{} (Section \ref{sec:awd_theory})
	\item Did you include complete proofs of all theoretical results?
    \answerYes{} (Section \ref{sec:awd_theory})
\end{enumerate}

\item If you ran experiments...
\begin{enumerate}
  \item Did you include the code, data, and instructions needed to reproduce the main experimental results (either in the supplemental material or as a URL)?
    \answerYes{} (Supplementary)
  \item Did you specify all the training details (e.g., data splits, hyperparameters, how they were chosen)?
    \answerYes{} (Section \ref{sec:exp})
	\item Did you report error bars (e.g., with respect to the random seed after running experiments multiple times)?
    \answerYes{} (Table \ref{tab-lm-results})
	\item Did you include the total amount of compute and the type of resources used (e.g., type of GPUs, internal cluster, or cloud provider)?
    \answerYes{} (Section \ref{sec:exp})
\end{enumerate}

\item If you are using existing assets (e.g., code, data, models) or curating/releasing new assets...
\begin{enumerate}
  \item If your work uses existing assets, did you cite the creators?
    \answerYes{} (Section \ref{sec:exp})
  \item Did you mention the license of the assets?
    \answerNA{}
  \item Did you include any new assets either in the supplemental material or as a URL?
    \answerNA{}
  \item Did you discuss whether and how consent was obtained from people whose data you're using/curating?
    \answerNA{}
  \item Did you discuss whether the data you are using/curating contains personally identifiable information or offensive content?
    \answerNA{}
\end{enumerate}

\item If you used crowdsourcing or conducted research with human subjects...
\begin{enumerate}
  \item Did you include the full text of instructions given to participants and screenshots, if applicable?
    \answerNA{}
  \item Did you describe any potential participant risks, with links to Institutional Review Board (IRB) approvals, if applicable?
    \answerNA{}
  \item Did you include the estimated hourly wage paid to participants and the total amount spent on participant compensation?
    \answerNA{}
\end{enumerate}

\end{enumerate}

\newpage

\appendix

\section*{Appendix}

\section{Double-LSTM} 
\label{sec:app_double-lstm}
\begin{wrapfigure}{R}{5cm}
    \vspace{-0.7cm}
    \includegraphics[width=0.95\linewidth]{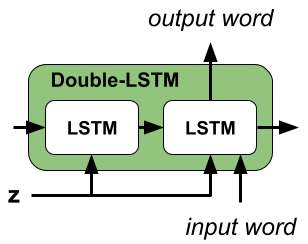}
    \captionof{figure}{\textbf{Double-LSTM.}}
    \label{fig-double-lstm}
\end{wrapfigure}
Inspired by the recent work \citep{dieng2019avoiding} 
that elucidates how \emph{skip-connections} promote higher latent information content, 
we introduce a simple-to-implement modification of a standard LSTM \citep{hochreiter1997long}.
Double-LSTM aims to promote the utilization of the latent variable $z$, whilst increasing the expressive power of the autoregressive decoder.
Double-LSTM consists of two LSTM units \citep{hochreiter1997long} as depicted in Figure~\ref{fig-double-lstm}.
The first LSTM unit is updated based on the latent variable $z$ and the previous hidden state $\bm{h}$.
The second LSTM unit is updated based on $z$, $\bm{h}$ and the input word embedding $\bm{w}$, which is subject to teacher forcing  and dropout.
The benefit of Double-LSTM is that it provides a two-branched skip connection to link $z$ with the output word.
The state update performed by the first LSTM is guaranteed to extract information from latent states and not from ground truth input.
In practice, Double-LSTM leads to performance improvements with little cost in terms of memory and computation time.

\section{Implementation Architecture} 
\label{sec:app_architecture}
\begin{figure*}[!h] 
\centering	
\includegraphics[width=\textwidth]{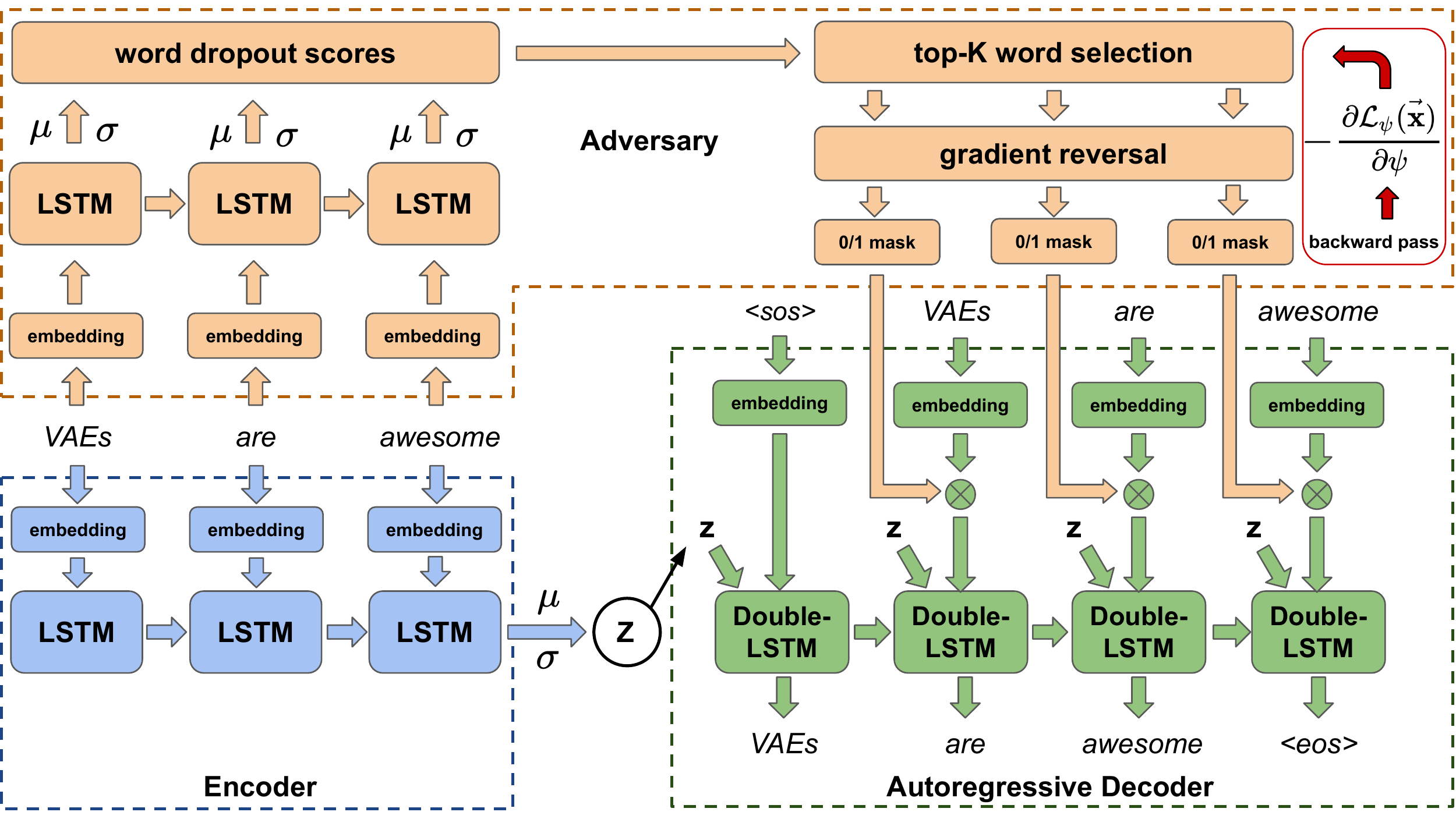}
\caption{\textbf{Architectural details of our proposed method:} Our variant of RNN-VAE \citep{bowman2016} consists of an LSTM-based sequence encoder {\color{cb}\textbf{(depicted in blue)}} and a Double-LSTM-based autoregressive sequence decoder {\color{celadon}\textbf{(depicted in green)}}. During MLE training, the ground truth sequence history is \emph{teacher-forced} to the autoregressive decoder which reduces the utilization of $z$. To regularize autoregressive decoding, the \emph{adversary} $\text{A}_{\psi}$ {\color{cadmiumorange}\textbf{(depicted in orange)}} obstracts VAE by \emph{learning to drop out} words (sequence elements) that VAE requires. Concretely, $\text{A}_{\psi}$ selectively masks different words at decoder's input, effectively performing input-conditioned word dropout. In the first stage, $\text{A}_{\psi}$ stochastically produces unnormalized \emph{word dropout scores} for each word in a sequence, using another LSTM that architecturally mirrors the one in the encoder. In the second stage, $\text{A}_{\psi}$ selects a subset of $K$ 'dropouts' using the differentiable \emph{top-$K$ word selection} module. Finally, to ensure that $\text{A}_{\psi}$ optimizes an objective that is an inverted version of the VAE objective, \emph{gradient reversal} \citep{ganin2016domain} is applied to negate the gradients in the backward pass.}
\label{fig-adversarial-dropout-appendix}
\end{figure*}

\section{Additional Quantitative Analysis}
\label{sec:app_qc}

\begin{table*}[h]
\centering
\begin{tabular}{l c c c}
 \toprule
 \bf   &  -ELBO $\downarrow$ & KL $\uparrow$ & MI $\uparrow$ \\
 \hline
 \\
 \textbf{Existing Sequence VAEs} & & & \\
Annealing \cite{bowman-etal-2015-large}    & 101.2 &  0.00 & -\\
 $\beta$ - VAE \cite{Higgins2017betaVAELB}    & 104.5 &  7.50 & 3.1\\
 SA \cite{kim2018semi}   & 102.6  & 1.23 & 0.7\\
   \vspace{0.2cm}
 Cyclical \cite{fu-etal-2019-cyclical}   & 103.1  & 3.48 & 1.8\\
 \hline
 \\
 \textbf{Our Sequence VAE} & & & \\
 unregularised   & 102.6 (0.3) & 1.1 (0.1) & 0.8 (0.4)\\
 + Word Dropout [0.4]   & 101.4 (0.3) & 5.1 (0.2) & 4.2 (0.3) \\
 + AWD [0.2]   & \textbf{99.7} (0.2) & 5.1 (0.1) & \textbf{4.3} (0.3) \\
 \bottomrule
 \\
 \end{tabular}
 \caption{\textbf{Results of text modeling on the PTB dataset} \cite{marcus-etal-1993-building}. Standard deviations are provided in the brackets.
    Squared bracket contains the dropout rate $DR$.
    ELBO -- evidence lower bound; KL - KL divergence;  MI -- mutual information.
 } 
 \label{PTB comparison}

\end{table*}

\begin{table}[h]
\centering
\begin{tabular}{l c c c}
 \toprule
 \bf   &  -ELBO $\downarrow$ & KL $\uparrow$ & MI $\uparrow$ \\
 \hline
 \\
 \textbf{Existing Sequence VAEs} & & & \\

 Annealing \cite{bowman-etal-2015-large}    & 33.08 &  1.42 & -\\
 Lagging \cite{he2018lagging}   & 32.95  & 1.42 & -\\
 Cyclical \cite{fu-etal-2019-cyclical}   & 34.32  & 3.63 & -\\
 FBP \cite{li2019surprisingly}   & 34.25  & \textbf{8.99} & -\\
 \hline
 \\
 \textbf{Our Sequence VAE} & & & \\
 unregularised   & 34.61 (0.1) & 0.06 (0.04)  & 0.8 (0.1) \\
 + Word Dropout [0.4]   & 33.82 (0.2) & 6.04 (0.6)  & 5.88 (0.6) \\
 + AWD [0.3]   & \textbf{32.66} (0.2) & 8.01 (0.7) & \textbf{7.22} (0.8) \\
 \bottomrule
 \\
 \end{tabular}
 \caption{\textbf{Results of text modeling on the SNLI dataset}. Standard deviations are provided in the brackets.
    Squared bracket contains the dropout rate $DR$.
    ELBO -- evidence lower bound; KL - KL divergence;  MI -- mutual information.
 } 
 \label{SNLI comparison}

\end{table}


\section{Definition of "posterior collapse"}
\label{sec-posterior-def}
We note that that the phenomena we address is often referred to as \textit{posterior collapse}, which could be misinterpreted as meaning that the posterior in fact \textit{collapses} to a single point (as in an MLE or MAP estimate), particularly since for VAEs, latent posteriors are typically more concentrated than the prior. As such, it might be less ambiguous to refer to the phenomenon as \textit{posterior dispersion} or \textit{posterior ignorance} or similar to better capture the fact that the posterior becomes diffuse and carries no information with respect to the data.

\section{Additional Qualitative Analysis}
\label{sec:app_qa}

This section provides additional qualitative experiments performed using the sequence VAE trained with the proposed adversarial training.

\begin{table}[h!]
    \centering
    {\def\arraystretch{1}\setlength{\tabcolsep}{3pt}
    \begin{tabular}{@{}l@{}}
    \toprule
    \textbf{I 'm not sure what all the hype is about . i 've been here a few times and it 's just ok . nothing} \\
    \textbf{special . I would n't go out of my way to come here .}\vspace{0.1cm}\\
    \emph{\quad I 've been here a few times and it 's always been good . the food is good , but the service is} \\
    \emph{\quad not so great .}\\
    \emph{\quad I 've been here a few times and it 's always been good . i 've had the chicken and waffles and} \\
    \emph{\quad the service was good .}\\
    \emph{\quad Great place to go for a quick bite to eat . the food is great and the service is great . i have} \\
    \emph{\quad been here a few times and have never been disappointed .}\vspace{0.1cm}\\
    \textbf{Great food , great service , and great service . i 've been} \textbf{here a few times and have never been} \\
    \textbf{disappointed}\\
    \bottomrule
    \end{tabular}}
    \vspace{0.2cm}
    \caption{\textbf{Sentence interpolation on the Yelp dataset.}}
\end{table}

\begin{table}[h!]
    \centering
    {\def\arraystretch{1}\setlength{\tabcolsep}{3pt}
    \begin{tabular}{@{}l@{}}
    \toprule
    <sos> do you think that you should be with someone you love ? if you do n't know what you are , \\then you should be happy . <eos> \\
    i am not sure if he is going to get a job . i am not ready to get married but i am not sure how to get \\
    him to pay for it . <eos> \\ 
    <sos> please help me with my homework ? i am a \_unk student and i need to know what the\\
    average salary of a school is in the us . i am looking for a website that has a list of the average \\
    salaries of students in the united states. <eos> \\
    <sos> what are some good websites to get free stuff on the net ? <eos> \\
    <sos> how do i become a better person ? i 'm shy , but i do n't know how to approach a guy . \\
    what should i do ? you should be able to be friends with someone who is not interested in you . \\ if you are shy.
    <sos> if you have a quadratic equation , what is the value of x ?  x = -1 <eos> \\
    <sos> how to get a \_unk visa ? i have a degree in psychology and i want to know what is the \\process  of getting a job in the us . if you are a \_unk , you can apply for a job . <eos> \\
    <sos> the u.s. has a nuclear power to stop the war in iraq ? what is the reason for the war ? \\ the war is over . <eos> \\
    <sos> question about jesus ? what is the name of the church that jesus is in the bible ? what is the \\name of the church ? <eos> \\
    <sos> do you think it is bad for you to have a cold sore ? i have a cold sore and i have a bad breath .\\ why do i have to pee ? it 's because it 's not a bad thing . <eos> \\
    <sos> is there any \_unk in the world ? yes , but there is no such thing as a soul mate . <eos> \\
    \bottomrule
    \end{tabular}}
    \vspace{0.2cm}
    \caption{\textbf{Unconditional sentence generation based on the Yahoo dataset.}}
\end{table}

\begin{table}[h!]
    \centering
    {\def\arraystretch{1}\setlength{\tabcolsep}{3pt}
    \begin{tabular}{@{}l@{}}
    \toprule
    \textbf{<sos> since the nfl has gone to the eight division format ? have three teams from teh same}\\
    \textbf{division made the playoffs in the same year not yet but could happen this year with the}\\
    \textbf{cowboys , giants , and redskins in the nfc east , or steelers , bengals}\\
    \textbf{, and ravens in the afc north .}\\
    \midrule
    <sos> the dog jumps into the air to catch a toy in its mouth . <eos> \\
    <sos> a young woman in a white shirt and black pants is playing with \\ 
    a young boy in a blue shirt . <eos> \\
    <sos> the person is flying a plane . <eos>\\
    <sos> the dogs have their owners in the air in front of a crowd of \\ 
    onlookers . <eos> \\{}
    <sos> the people are participating in an operation . <eos> \\
    <sos> a woman with black hair is standing in a puddle . <eos> \\
    <sos> a young woman is riding a bike in front of a group of people in \\ 
    a red dress . \\
    <sos> people are holding up their signs in their hands . <eos> \\
    \bottomrule
    \end{tabular}}
    \vspace{0.2cm}
    \caption{\textbf{Neighborhood exploration based on the Yahoo dataset.} The original sentence taken from the Yahoo dataset \emph{(on top)} was used to infer the parameters of the posterior. We then sampled from the posterior and decoded the sentences multiple times.}
\end{table}

\end{document}